\documentclass[lettersize,journal]{IEEEtran}
\usepackage{amsmath,amsfonts}
\usepackage{algorithm}
\usepackage{array}
\usepackage[caption=false,font=normalsize,labelfont=sf,textfont=sf]{subfig}
\usepackage{textcomp}
\usepackage{stfloats}
\usepackage{url}
\usepackage{verbatim}
\usepackage{graphicx}
\usepackage{cite}
\usepackage{booktabs}
\usepackage{lscape}
\usepackage{soul, color, xcolor}
\usepackage{multirow}
\usepackage{epstopdf}
\usepackage{algorithm}
\usepackage{algorithmicx}
\usepackage{algpseudocode}
\usepackage{float} 
\usepackage{subfloat}
\usepackage{caption}
\usepackage[english]{babel}
\usepackage{blindtext}
\usepackage{tabularx}
\usepackage{threeparttable} % For table notes
\usepackage{subcaption}
\usepackage{comment}
\usepackage{makecell}
\usepackage{booktabs}
\usepackage{array}
\usepackage{multirow}
\usepackage{hyperref}
\hypersetup{hypertex=true,
colorlinks=true,
linkcolor=blue,
anchorcolor=blue,
citecolor=blue}

\soulregister\cite7

\soulregister\eqref7

\begin{document}

\title{OMUDA: Omni-level Masking for Unsupervised Domain Adaptation in Semantic Segmentation}

\author{Yang Ou, \IEEEmembership{Member, IEEE}, Xiongwei Zhao$^{\dagger}$, \IEEEmembership{Member, IEEE}, Xinye Yang, Yihan Wang, \\  Yicheng Di, Rong Yuan, Xieyuanli Chen, \IEEEmembership{Member, IEEE}, Xu Zhu, \IEEEmembership{Senior Member, IEEE}

 % <-this % stops a space
% \thanks{This work was supported in part by Marine Economy Development Project of Guangdong Province under Grant GDNRC [2020]014; in part by the School-Level Scientific and Technological Research Initiative: Key Research Area Support Program (Science and Technology) under Grant SZIIT2024KJ047. (Corresponding author: Xiongwei Zhao)}

% <-this % stops a space
\thanks{Yang Ou and Rong Yuan are with the School of Mechanical Engineering, Chengdu University, Chengdu 610106, China (e-mail: ouyang@my.swjtu.edu.cn, Yuanrong@cdu.edu.cn).}
\thanks{Xiongwei Zhao and Xu Zhu are with the School of Information Science and Technology, Harbin Institute of Technology, Shenzhen 518055, China (email: xwzhao@stu.hit.edu.cn, xuzhu@ieee.org)}
\thanks{Xinye Yang is with the School of Computer Science and Informatics, Cardiff University, Cardiff CF24 4AG, United Kingdom (email: yangxinye333@gmail.com)}
\thanks{Yihan Wang is with the School of Mechanical Engineering, Shouthwest Jiaotong University, Chengdu 610031, China (email: wyh123@my.swjtu.edu.cn)}
\thanks{Yicheng Di is with the School of Artificial Intelligence and Computer Science, Jiangnan University, Wuxi, China (email: diyicheng@stu.jiangnan.edu.cn)}
\thanks{Xieyuanli Chen is with the College of Intelligence Science and Technology, National University of Defense Technology, Changsha 410073, China (email: xieyuanli.chen@nudt.edu.cn).}

% \thanks{The datasets and code will be made publicly available at {\textcolor{red}{\textit{\url{https://github.com/Grandzxw/ITDNet}}}}}
\thanks{$\dagger$ indicates Corresponding author.}
% \thanks{Corresponding author: Xiongwei Zhao.}
}

\maketitle
\begin{abstract}
Unsupervised domain adaptation (UDA) enables semantic segmentation models to generalize from a labeled source domain to an unlabeled target domain. However, existing UDA methods still struggle to bridge the domain gap due to cross-domain contextual ambiguity, inconsistent feature representations, and class-wise pseudo-label noise. To address these challenges, we propose Omni-level Masking for Unsupervised Domain Adaptation (OMUDA), a unified framework that introduces hierarchical masking strategies across distinct representation levels. Specifically, OMUDA comprises: 1) a Context-Aware Masking (CAM) strategy that adaptively distinguishes foreground from background to balance global context and local details; 2) a Feature Distillation Masking (FDM) strategy that enhances robust and consistent feature learning through knowledge transfer from pre-trained models; and 3) a Class Decoupling Masking (CDM) strategy that mitigates the impact of noisy pseudo-labels by explicitly modeling class-wise uncertainty. This hierarchical masking paradigm effectively reduces the domain shift at the contextual, representational, and categorical levels, providing a unified solution beyond existing approaches. Extensive experiments on multiple challenging cross-domain semantic segmentation benchmarks validate the effectiveness of OMUDA. Notably, on the SYNTHIA$\rightarrow$Cityscapes and GTA5$\rightarrow$Cityscapes tasks, OMUDA can be seamlessly integrated into existing UDA methods and consistently achieving state-of-the-art results with an average improvement of 7\%.

% feature consistency for place recognition
% Dual-Domain Mixer
% Semantic-Aware Generator

\end{abstract}


% \begin{figure}[!t]
% \centerline{\includegraphics[width=\columnwidth]{fig1.pdf}}
% \caption{Visualization of semantic 3D points from the same LiDAR frame across four weather conditions in our proposed Weather-KITTI dataset. Weather noises are marked in \textcolor{red}{red}.} 
% \vspace{-1.2em}
% \label{distrubution}
% \end{figure}
%% Separate Training

\section{Introduction}
\label{Section I}
\IEEEPARstart{S}{emantic} segmentation, a fundamental task in computer vision, aims to assign a semantic category to each pixel in an image. Recent advancements in deep learning-based methods have shown remarkable success in this field~\cite{guo2024contrastive,yu2025rethinking,triplemixer,2025iterative,kuang2025reslpr,yang2025splatpose}. However, such success heavily depends on large-scale annotated training data, whose collection is both labor-intensive and time-consuming~\cite{vijai2025artificial,zhou2025aeseg,ahn2024style}. To alleviate this annotation burden, researchers have explored leveraging synthetic data to train semantic segmentation models~\cite{loiseau2024reliability,qian2024maskfactory,tang2025training}. Although synthetic data offers a cost-effective alternative to manual annotation, models trained solely on synthetic images often generalize poorly to real-world domains due to the substantial domain gap between synthetic and real data distributions.

\begin{figure*}[!t]
\centering
\includegraphics[width=1.0\linewidth]{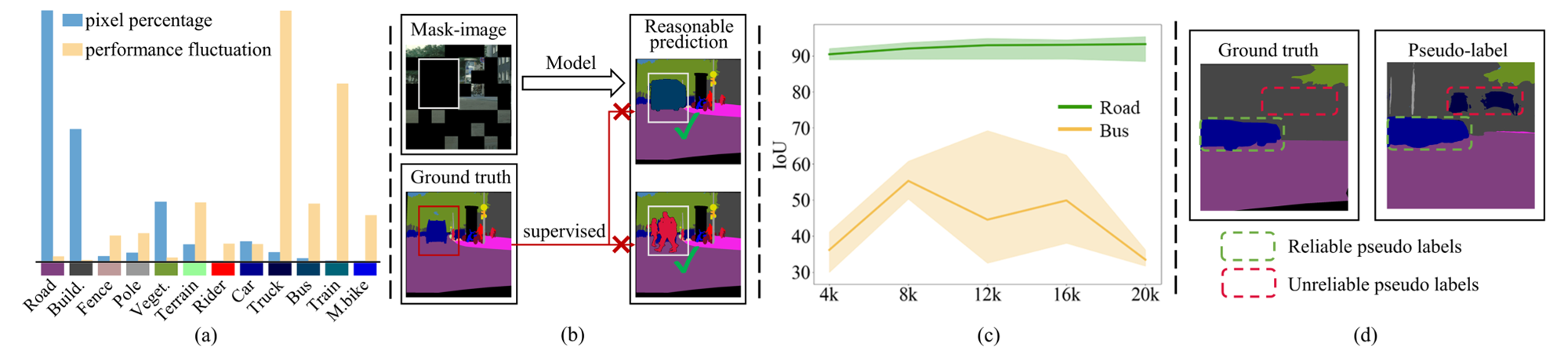}

\caption{Key inconsistency issues in existing UDA methods\cite{du2022learning, zhou2022context, yang2024micdrop}. (a) Class performance fluctuation is inversely correlated with pixel frequency, with foreground classes showing higher instability. (b) Uniform masking disrupts critical background context and misguides supervision of foreground regions. (c) Background features remain stable during training, whereas foreground features easily drift and overfit. (d) Pseudo-label reliability varies across classes, with rare or complex categories exhibiting more errors.}
\label{fig:mov_method}
\vspace{-1.0em}
\end{figure*}

To address the above limitations, unsupervised domain adaptation (UDA) methods~\cite{udasurvey2,udasurvey1} have been developed to transfer pixel-level knowledge from a labeled source domain to an unlabeled target domain. However, existing UDA methods~\cite{mic,du2024domain,sam4udass,micdrop} typically adopt a uniform strategy for all classes, ignoring the substantial differences between foreground and background categories. Background classes dominate the pixel distribution and largely determine the scene type, thereby providing rich contextual cues. In contrast, foreground classes have well-defined shapes and object boundaries but contribute less to global context. Despite their distinct characteristics, both background and foreground classes are crucial for comprehensive scene understanding. However, as illustrated in Fig.~\ref{fig:mov_method}, empirical results of existing UDA methods show that they behave quite differently during training: (1) Compared with background classes, foreground categories (e.g., truck, train) exhibit much larger performance fluctuations across trials, as shown in Fig.~\ref{fig:mov_method}(a), and form less stable contextual and spatial structures, whereas background classes maintain coherent layouts and strong contextual dependencies, as illustrated in Fig.~\ref{fig:mov_method}(b). (2) foreground classes are more vulnerable to early performance degradation due to overfitting, given their limited and diverse samples, as demonstrated in Fig.~\ref{fig:mov_method}(c); (3) large disparities in per-class accuracy lead to heterogeneous pseudo-label reliability across categories, as illustrated in Fig.~\ref{fig:mov_method}(d). These issues limit the performance of existing UDA methods on semantic segmentation, and naturally motivate the design of of diversified masking strategies that treat different semantic categories.

To address the aforementioned challenges, we propose Omni-level Masking for Unsupervised Domain Adaptation (OMUDA) for semantic segmentation, where Omni-level indicates that the proposed framework operates jointly at the contextual, representational, and categorical levels of adaptation. First, we introduce a Context-Aware Masking (CAM) strategy that combines class-aware sampling with distinct masking for foreground and background regions. CAM assigns different sampling weights to each class according to its pixel occupancy, and then applies coarse-grained masking to background regions to preserve global context and fine-grained masking to foreground regions to retain object-level details. Next, to reduce early overfitting and feature instability in foreground classes, we design the Feature Distillation Masking (FDM) strategy, which leverages generalized features from a pre-trained model to constrain foreground predictions during training. This encourages the model to learn more stable and consistent representations for domain-sensitive classes, particularly rare or small objects. Finally, to handle the varying reliability of pseudo-labels across categories, we propose the Class Decoupling Masking (CDM) strategy. CDM introduces a class-wise weighting mechanism that accounts for differences in confidence and accuracy, down-weighting unreliable pseudo-labels from difficult-to-adapt classes and thereby mitigating their negative impact during training. By integrating these three complementary strategies, OMUDA provides a robust and generalizable solution for UDA in semantic segmentation, with broad applicability to domain-shift scenarios in computer vision. Our contributions in this paper are summarized as follows:
\begin{itemize}
    \item We propose Omni-level Masking for Unsupervised Domain Adaptation (OMUDA) for semantic segmentation, which introduces a hierarchical masking framework to enhance cross-domain knowledge transfer and alleviate performance degradation caused by domain shifts.
    \item We develop three complementary masking strategies acting at the contextual, representational, and categorical levels, respectively, effectively addressing foreground–background discrepancies, instability in foreground representations, and class-wise pseudo-label inconsistency within a unified framework.
    \item Extensive experiments on standard UDA benchmarks demonstrate that OMUDA can be seamlessly integrated into existing methods, consistently boosting their performance and achieving state-of-the-art results for semantic segmentation under domain shift.
\end{itemize}

The remainder of the paper is organized as follows. Section~\ref{Section II} reviews related work. Section~\ref{Section III} describes the proposed OMUDA framework and its three masking strategies. Section~\ref{Section IV} presents experimental results and ablation studies. Finally, Section~\ref{Section V} concludes the paper.

\section{Related Work}
\label{Section II}
\subsection{Semantic Segmentation}
\label{Section II.A}
Recent advances in deep learning have significantly enhanced the accuracy and efficiency of semantic segmentation models\cite{li2024transformer,zhu2024survey, ran2024pseudo}. The authors of \cite{hu2021} and \cite{wang2021} proposed fully supervised methods that exploit global pixel-level relationships by extracting features across multiple images to regularize the segmentation embedding space. The authors of \cite{peng2022} introduced a unified attention fusion module, which leverages both spatial and channel attention to compute adaptive weights for feature fusion, enhancing representational capacity while maintaining lightweight efficiency. The authors of \cite{sung2024} proposed a contrastive learning-based semantic segmentation method that allows to capture local/global contexts and comprehend their relationships. However, the success of these methods often relies on large-scale annotated datasets, which are expensive and labor-intensive to obtain in real-world scenarios. To alleviate this issue, synthetic datasets~\cite{loiseau2024reliability,silva2025exploring,triplemixer} with dense annotations are commonly used as the labeled source domain. However, the domain gap between synthetic and real-world data, caused by differences in lighting, weather, sensor configurations, or geographic environments, still remains a major challenge.

\subsection{Unsupervised Domain Adaptation (UDA)}
\label{Section II.B}
Unsupervised domain adaptation (UDA) for semantic segmentation has been proposed to transfer knowledge from a labeled source domain to an unlabeled target domain, without requiring manual annotations. Existing UDA approaches for semantic segmentation can be broadly categorized into two main approaches: adversarial learning and self-learning~\cite{fang2024source,hu2023multi,2023survey}. Adversarial training aims to reduce the distribution gap between source and target domains by aligning feature representations~\cite{tian2022adt,yang2022adt}. Following the success of generative adversarial networks (GANs), these approaches typically employ a domain discriminator to distinguish source from target features, while the segmentation network is trained to generate domain-invariant representations~\cite{vu2019advent,wang2021interbn}. More recently, class-conditional adversarial approaches focus on aligning features within the same category rather than enforcing strict global alignment, as effective adaptation can be achieved when features are well separated across classes~\cite{luo2019taking,luo2021category}. Despite their progress, adversarial training approaches~\cite{zheng2022adaptive} often suffer from instability, mode collapse, and sensitivity to discriminator design, leading to suboptimal segmentation performance. In contrast, self-training approaches~\cite{chen2023pipa,li2025unsupervised,zhao2024unsupervised} have recently become the dominant paradigm. They typically use a model trained on the labeled source domain to generate pseudo-labels for unlabeled target images, which are then exploited for iterative re-training. However, due to the substantial distribution discrepancy between source and target domains, the generated pseudo-labels inevitably contain noise, leading to error accumulation and potential performance degradation.

\begin{figure*}
\centering
\includegraphics[width=1.0\linewidth]{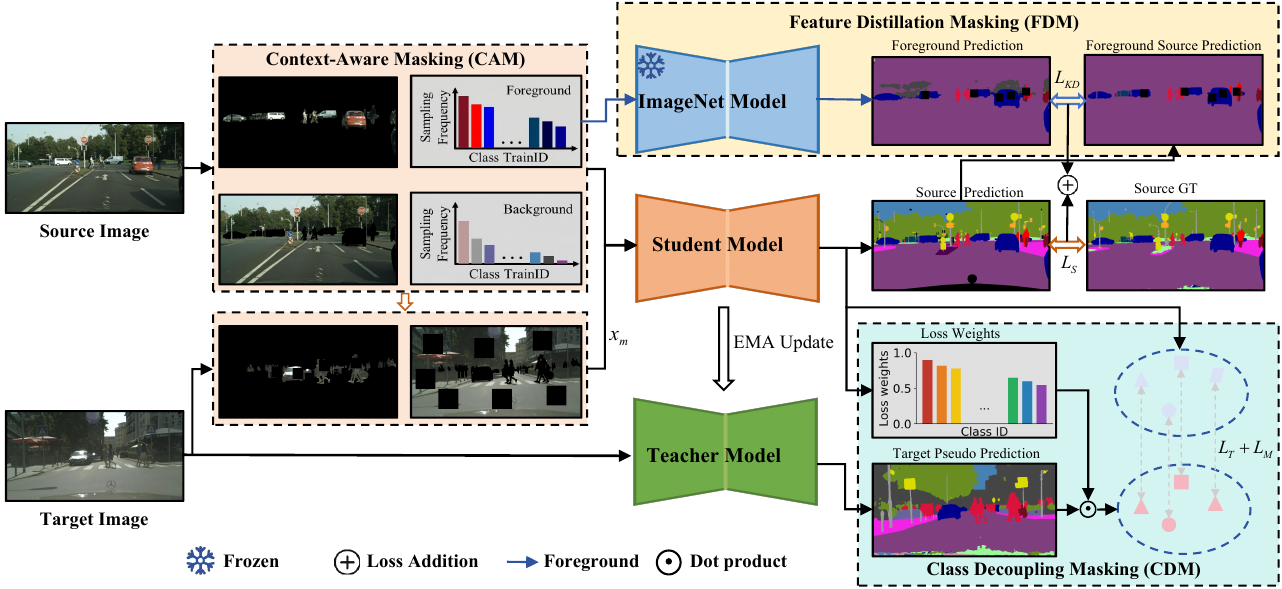}
\caption{Overview of the proposed OMUDA framework. The framework integrates three modules: CAM for class-aware sampling and masking, FDM for foreground feature distillation using a pretrained ImageNet model, and CDM for class-wise reliability weighting and pseudo-label refinement. A student–teacher architecture with EMA updates enhances stability and effectiveness in unsupervised domain adaptation for semantic segmentation.}
\label{fig:framework}
\vspace{-1.0em}
\end{figure*}

\subsection{Enhancing Self-training Approaches in UDA}
\label{Section II.C}
To mitigate the issue of pseudo-label drift in self-training, various regularization strategies have been proposed. A common line of work focuses on confidence-based filtering, where only high-confidence pseudo-labels are retained for retraining~\cite{wu2021dannet,zou2018unsupervised,zou2019confidence}. Beyond simple thresholding, prototype-based methods introduce sample-wise pseudo-label correction~\cite{zhang2021prototypical,zhao2022source}, leveraging class prototypes to refine noisy predictions and improve label reliability. In parallel, other studies have incorporated additional learning constraints to stabilize training. Consistency regularization encourages the model to produce invariant predictions under diverse perturbations of the same input~\cite{araslanov2021self,wang2022exploring}, while contrastive learning enhances feature discriminability by explicitly separating positive and negative pairs~\cite{lee2022bi,liu2021domain}. Moreover, cross-domain mixup~\cite{tranheden2021dacs,zhang2025unified} has been proposed as a data augmentation technique to create interpolated samples between the source and target domains, thereby smoothing the transition between domains. Uncertainty estimation~\cite{cheng2021dual,lu2025uncertainty,yin2024uncertainty,zheng2021rectifying} has also been utilized to assess the reliability of pseudo-labels and guide the adaptation process. However, a key limitation of existing approaches is that they typically treat all classes uniformly, overlooking the distinct adaptation behaviors of foreground and background categories. This one-size-fits-all paradigm is often suboptimal, as semantic classes exhibit varying degrees of domain shift and adaptation difficulty.

%\section{Proposed Omni Mask Domain Adaption scheme}
\section{Methodology}
\label{Section III}
In this section, we first present the problem statement of UDA semantic segmentation, followed by a detailed introduction of our proposed OMUDA framework. The Pipeline overview of our OMUDA framework is shown in Fig. \ref{fig:framework}. As illustrated in Fig. \ref{fig:framework}, OMUDA comprises three key masking strategies: Context-Aware Masking (CAM), Feature Distillation Masking (FDM), and Class Decoupling Masking (CDM). The details of each component are described below.

\subsection{Problem Statement}
\label{Section III.A}
Unsupervised domain adaptation (UDA) semantic segmentation contains two datasets: a labeled source dataset $\mathbb{D}_s=\{(x_s^{n},y_s^{n})\}_{n=1}^{N_s}$ and an unlabeled target dataset $\mathbb{D}_t=\{x_t^{m}\}_{m=1}^{N_t}$. Each image $x \in \mathbb{R}^{H\times W}$ has height $H$ and width $W$, and the corresponding source annotation $y_s^{n}\in\{0,1\}^{H\times W\times K}$ is a one-hot label map over $K$ semantic classes. Although the sizes $N_s$ and $N_t$ may differ, both domains share the same label space of $K$ classes. The goal of UDA semantic segmentation is to learn a domain-invariant segmentation network $\varphi$ that performs well on the target domain by jointly exploiting $\mathbb{D}_s$ and $\mathbb{D}_t$.

In practice, a network $\varphi$ trained only on $\mathbb{D}_s$ often generalizes poorly to $\mathbb{D}_t$ due to the domain gap. To mitigate this issue, self-training UDA methods generate pseudo labels $\hat{y}_t$ for target images and optimize a combined cross-entropy (CE) loss on source and target data. For a source image $x_s$ with label $y_s$ and a target image $x_t$ with pseudo label $\hat{y}_t$, the CE loss can be written as:
\begin{equation}
    \mathcal{L}_{ce} \!=\! -\!\sum_{i=1}^{H\times W}\! \sum_{k=1}^K y^{(i,k)}_s \log	\varphi(x_{s})^{(i,k)}\!+\hat{y}^{(i,k)}_t \log\varphi(x_{t})^{(i,k)}
\end{equation}
where $\hat{y}^{(i,k)}_t=\arg \max_{k} \phi(x_{t})^{(i,k)}$ is the pseudo label predicted by the teacher network $\phi$. The teacher network $\phi$ is not updated by backpropagation, but instead by the exponentially moving average (EMA) of the student segmentation network $\varphi$ after each step $j$:
\begin{equation}
    \phi_{j+1}=\alpha\phi_j+(1-\alpha)\varphi_j
\end{equation}
where $\alpha \in [0,1)$ is the EMA decay factor.

\subsection{Context-Aware Masking (CAM)}
\label{Section III.B}
Substantial variations in performance fluctuation are observed across different classes. As illustrated in Fig.\ref{fig:mov_method}(a), the fluctuation magnitude is inversely correlated with the percentage of pixels belonging to each class. Foreground classes, which typically occupy smaller regions and contain fewer pixels, exhibit significantly higher performance volatility than background classes. In this work, we follow the common distinction in semantic segmentation between object and background categories: 
foreground classes refer to object categories with clear instance boundaries (e.g., person, rider, car), 
whereas background classes correspond to large, layout-defining regions (e.g., road, building, sky). 
This observation suggests that class imbalance is a major contributing factor to unstable learning. Motivated by this observation, we assign class-specific sampling weights based on pixel occupancy. We first compute the pixel frequency $f_k$ of class $k$ as the proportion of pixels labeled as class $k$ over the source domain:
\begin{equation}
\begin{aligned}
f_k=\frac{\sum_{n=1}^{N_s} \sum_{i=1}^{H \times W}{\mathbb{I}\left[{y}_s^{(i, n)}=k \right]}}{N_s \times H \times W}
\end{aligned}
\end{equation}

To better accommodate the distinct characteristics of foreground and background classes, we decouple them and compute their sampling probabilities separately:
\begin{equation}
\begin{aligned}
& P_b(k)=\frac{e^{\left(1-f_k\right) / T_b}}{\sum_{k^{\prime}=1}^{K_b} e^{\left(1-f_k^{\prime}\right) / T_b}}
\end{aligned}
\end{equation}
\begin{equation}
\begin{aligned}
& P_f(k)=\frac{e^{\left(1-f_k\right) / T_f}}{\sum_{k^{\prime}=1}^{K_f} e^{\left(1-f_k^{\prime}\right) / T_f}}
\end{aligned}
\end{equation}
where $T_b$ and $T_f$ control the smoothness of the sampling distribution for background and foreground classes, respectively, while $K_b$ and $K_f$ denote the number of background and foreground classes. The final sampling distribution is given by $P = \{P_b, P_f\}$. In all experiments, we set $T_b = 1.0$ and $T_f = 0.7$ to provide a moderate re-balancing effect for background classes while more strongly emphasizing rare foreground categories.

On the other hand, foreground classes typically lack strong contextual dependencies among themselves. In UDA semantic segmentation, this further complicates representation learning because standard masking strategies (e.g., random or grid-based masking) treat all regions uniformly and overlook the distinct roles of foreground and background. As shown in Fig.~\ref{fig:mov_method}(b), such indiscriminate masking can disrupt essential background structures (e.g., road or sky), which provide key contextual cues, leading to semantic inconsistencies such as misclassifying a “car’’ as a “person’’. To address this issue, we obtain pseudo labels $\hat{y}_t$ from the teacher network $\phi$ and partition each image into foreground and background regions based on the predefined class sets $K_f$ and $K_b$. We then apply a coarse-grained mask $\mathcal{M}_b$ to background regions and a fine-grained mask $\mathcal{M}_f$ to foreground regions:
\begin{equation}
\begin{aligned}
    x_{t_b} &= \mathcal{M}_b \odot x_t, \\
    x_{t_f} &= \mathcal{M}_f \odot x_t,
\end{aligned}
\end{equation}
where $\odot$ denotes element-wise multiplication. The final masked image $x_m$ is obtained by combining $x_{t_b}$ and $x_{t_f}$ based on the pseudo labels:
\begin{equation}
\begin{aligned}
    x_m = x_{t_b} \odot {\mathbb{I}\left[\hat{y}^{(i,k)}_t \in K_b \right]}+x_{t_f} \odot {\mathbb{I}\left[\hat{y}^{(i,k)}_t \in K_f \right]}
\end{aligned}
\end{equation} 

The CAM strategy first performs class-aware sampling and then applies distinct masking patterns to background and foreground regions, enabling the model to better capture their differing characteristics while reducing its over-reliance on background context during training. Unlike DACS\cite{tranheden2021dacs}, which adopts a uniform class-mixing scheme for all categories, CAM explicitly decouples foreground and background classes and tailors the masking process to their semantic roles in the scene.

\subsection{Feature Distillation Masking (FDM)}
\label{Section III.C}
During the training process, the model distinguishes classes by learning discriminative features from the input data. However, as illustrated in Fig.~\ref{fig:mov_method}(c), background features tend to be more stable, while foreground features exhibit greater variability and are more susceptible to overfitting. This instability comes mainly from two factors. First, foreground classes contain far fewer instances than background classes, limiting the data available for learning robust representations. Second, foreground categories often display more complex and diverse visual appearances, making it difficult for the model to learn generalizable features. Such unstable foreground representations ultimately hinder semantic segmentation performance. 

To mitigate this issue, we leverage a pretrained feature extractor $E_{pre}$, trained on a large-scale dataset (e.g., ImageNet), to provide stable foreground feature priors. For each input, we obtain feature maps from both $E_{pre}$ and the student model $\varphi$'s neck module $E_{neck}$, and introduce a feature distillation loss on foreground regions rather than relying solely on noisy pseudo-label supervision. Specifically, we distill both the feature distance and the angular relationship between foreground samples, which encourages $E_{neck}$ to preserve the similarity structure and geometric configuration of foreground embeddings. This regularization stabilizes foreground representations and effectively alleviates early overfitting and feature collapse. First, we compute the feature distance between different samples as:

\begin{equation}
\begin{aligned}
& d_{pre}^{(i,j)}=\frac{\left\|\left(f_{pre}^i-f_{pre}^j\right)M\right\|_2}{\sum_{\left(x_i, x_j\right) \in \mathbb{D}_s}\left\|\left(f_{pre}^i-f_{pre}^j\right)M\right\|_2}\\
& \text{with} \; f_{pre}^i=E_{pre}(x_s^i),\; M={\mathbb{I}\left[y^{(i,k)}_s \in K_f \right]}
\end{aligned}
\end{equation}
where $M$ is a mask that indicates the foreground classes in the image. Similarly, we compute $d_{neck}^{(i,j)}$ and the distance distillation loss function is defined as follows:
\begin{equation}
\begin{aligned}
& \mathcal{L}_{\text {D}}=\sum_{\left(x_i, x_j\right) \in \mathbb{D}_s} L_1(d_{pre}^{(i,j)}-d_{neck}^{(i,j)}) \\
& \text{with} \; L_1(x)= \begin{cases}0.5 x^2 & \text { if }|x|<1 \\ |x|-0.5 & \text { otherwise }\end{cases}
\end{aligned}
\end{equation}

The feature angle between different samples is calculated as follows:
\begin{equation}
\begin{aligned}
a_{pre}^{i, j, k}=\langle\frac{\left(f_{pre}^i-f_{pre}^j\right)M}{\|(f_{pre}^i-f_{pre}^j)M\|_2},\frac{\left(f_{pre}^i-f_{pre}^k\right)M}{\left\|\left(f_{pre}^i-f_{pre}^k\right)M\right\|_2}\rangle
\end{aligned}
\end{equation}
where $\langle\cdot\rangle$ is the cosine similarity, and the angular distillation loss function is:
\begin{equation}
\begin{aligned}
\mathcal{L}_{A}=\sum_{\left(x_i, x_j, x_k\right) \in \mathbb{D}_s}L_1(a_{pre}^{i, j, k}-a_{neck}^{i, j, k})
\end{aligned}
\end{equation}

The overall KD loss function is
\begin{equation}
\begin{aligned}
\mathcal{L}_{KD}=\mathcal{L}_{D}+\mathcal{L}_{A}
\end{aligned}
\end{equation}

Ultimately, $\mathcal{L}_{KD}$ is added to the final loss function via a weighting factor $\lambda_{KD}$. The FDM strategy integrates the mask-based KD loss $\mathcal{L}_{KD}$ into the training process, enhancing the student model’s ability to generate accurate masks and improving its segmentation and localization performance. This leads to more reliable and precise semantic segmentation results.

\subsection{Class Decoupling Masking (CDM)}
\label{Section III.D}
Fig.~\ref{fig:mov_method}(d) further shows that pseudo-label reliability varies substantially across classes. This variability stems from factors such as domain shift, distribution mismatch, and model uncertainty. Some categories exhibit higher prediction uncertainty due to appearance variations, occlusion, or inherent semantic ambiguity, which results in more labeling errors. Consequently, pseudo-labels for these classes tend to be less reliable than those for more stable categories. To address this issue, we introduce a weighting strategy that assigns higher weights to the pseudo-label errors of difficult-to-adapt classes. This mitigates their adverse influence during training, as such classes are more susceptible to domain shift. We begin by normalizing the class-wise loss. The normalized loss $\mathcal{L}_{norm}$ is obtained by summing over the spatial dimensions $(H \times W)$ and the classes $(K)$, while using the indicator function $\mathbb{I}\left[\hat{y}^{(i)}=k\right]$ to ensure that only the loss corresponding to the predicted class is counted. The formulation is given by:
\begin{equation}
\mathcal{L}_{norm}=\sum_{i=1}^{H \times W} \sum_{k=1}^K \frac{1}{\mathbb{I}\left[\hat{y}^{(i)}=k\right]} \hat{y}^{(i, k)} \log	\varphi(x)^{(i,k)} 
\end{equation}
\indent Next, we take into account the varying difficulty of migration across classes. We calculate a weighting factor $\beta_k$ for each class $k$, which quantifies the uncertainty of each training sample. The weighting factor is computed by summing over the spatial dimensions $(H \times W)$ and the samples $(N_t)$, and using the indicator function to determine if the predicted label $p_k^{(i,n)}$ matches the pseudo-label $\hat{y}_k^{(i, n)}$. The equation for $\beta_k$ is as follows:
\begin{equation}
\begin{aligned}
\beta_k=\sum_{i=1}^{H \times W}\sum_{n=1}^{N_t} \left(1-\frac{\mathbb{I}\left[p_k^{(i,n)}=\hat{y}_{k=1}^{(i, n)}\right]}{\mathbb{I}\left[\hat{y}_k^{(i, n)}=c\right]}\right) 
\end{aligned}
\end{equation}
\indent Finally, we incorporate the class weighting into the loss function. The normalized loss $\mathcal{L}_{norm}$ is multiplied by the weighting factor $\beta_k$, and the indicator function $\mathbb{I}\left[\hat{y}^{(i, k)}=j\right]$ is used to ensure that only the loss for the correct class is considered. The equation for the class weighting loss function is as follows:
\begin{equation}
\begin{aligned}
\mathcal{L}_{w\_norm}=\sum_{i=1}^{H \times W} \sum_{k=1}^K \frac{\beta_k}{\mathbb{I}\left[\hat{y}^{(i)}=k\right]} \hat{y}^{(i, k)} \log	\varphi(x)^{(i,k)} 
\end{aligned}
\end{equation}

\subsection{Loss Function}
\label{Section III.D}
In our framework, the student network $\varphi$ is trained with three types of supervision: 
fully supervised losses on the labeled source domain, feature distillation on source foreground regions, 
and pseudo-label based losses on the unlabeled target domain. 
On the source domain, we use the standard cross-entropy loss with ground-truth annotations, 
denoted as $\mathcal{L}_{S}$, and the foreground feature distillation loss $\mathcal{L}_{KD}$ between the student neck $E_{neck}$ and the pretrained feature extractor $E_{pre}$. 
On the target domain, we employ the class-weighted normalized loss $\mathcal{L}_{w\_norm}$, using teacher-generated pseudo-labels as supervision, 
to obtain the loss on original target images $\mathcal{L}_{T}$ and on their CAM-generated masked loss $\mathcal{L}_{M}$. 
Within the CDM module, $\mathcal{L}_{w\_norm}$ reweights classes according to their pseudo-label reliability, 
so that noisy and hard-to-adapt categories do not dominate the optimization. The overall training objective is:
\begin{equation}
    \mathcal{L}_{{Total}}
    = \mathcal{L}_{S}
    + \lambda_{KD} \mathcal{L}_{KD}
    + \mathcal{L}_{T}
    + \mathcal{L}_{M},
\end{equation}
where $\lambda_{KD}$ is the a weighting factor of KD loss, and its effect is further analyzed in the ablation studies in Section~\ref{Section IV}.

\section{Experimental results}
\label{Section IV}
\subsection{Datasets and Implementation Details}
\label{Section IV.A}
We conduct our experimental evaluation on two widely adopted UDA benchmarks. The first uses synthetic source images from the GTA5 dataset\cite{richter2016playing}, containing 24,966 images (1914×1052). The second employs SYNTHIA\cite{ros2016synthia} as the source domain, which provides 9,400 images (1280×760). In both benchmarks, the target domain is Cityscapes\cite{cordts2016cityscapes}, consisting of 500 validation images (2048×1024). Following prior work\cite{tranheden2021dacs,hoyer2022daformer,du2022learning}, we report mIoU over the 19 classes shared by GTA5→Cityscapes and the 16 classes shared by SYNTHIA→Cityscapes. We use DAFormer\cite{hoyer2022daformer} as our baseline model, as it achieves a strong performance at high training and inference speed. For pretrained feature extraction, we adopt MiT-B5 \cite{xie2021segformer} pretrained on ImageNet\cite{deng2009imagenet}.

We adopt AdamW\cite{loshchilov2017decoupled} with betas (0.9, 0.999) as the optimizer, a learning rate of $6 \times 10^{-5}$ for the encoder and $6 \times 10^{-4}$ for the decoder, a linear learning rate warmup\cite{goyal2017accurate} of 1.5k iterations, a weight decay of 0.01. Following DAFormer\cite{hoyer2022daformer}, we resize target images to 1024 × 512 and source images to 1280 × 720, with a random crop size of 512 × 512. The model is trained for 40k iterations with the batch size of 3. All experiments are conducted on one NVIDIA GeForce RTX 3090 and implemented with the PyTorch framework.

\begin{table*}[t]
\centering
\caption{Quantitative comparison with previous UDA methods on GTA5$\to$Cityscapes benchmark. The best results are highlighted in bold while the second best results are shown by underline.}
\label{tab1}
\setlength{\tabcolsep}{1pt}
\renewcommand{\arraystretch}{1}
\resizebox{\textwidth}{!}{%
\begin{tabular}{l|ccccccccccccccccccc|c}
\bottomrule
 & \rotatebox{45}{Road} & \rotatebox{45}{S.walk} & \rotatebox{45}{Build.} & \rotatebox{45}{Wall} & \rotatebox{45}{Fence} & \rotatebox{45}{Pole} & \rotatebox{45}{Tr.Light} & \rotatebox{45}{Sign} & \rotatebox{45}{Veget.} & \rotatebox{45}{Terrain} & \rotatebox{45}{Sky} & \rotatebox{45}{Person} & \rotatebox{45}{Rider} & \rotatebox{45}{Car} & \rotatebox{45}{Truck} & \rotatebox{45}{Bus} & \rotatebox{45}{Train} & \rotatebox{45}{M.bike} & \rotatebox{45}{Bike} & mIoU\\
\hline
\multicolumn{21}{c}{GTA5 $\to$ Cityscapes} \\
\hline
% AdaptSeg \cite{tsai2018learning}
% & 86.5 & 36.0    & 79.9     & 23.4 & 23.3  & 23.9 & 35.2  & 14.8 & 83.4       & 33.3    & 75.6 & 58.5   & 27.6  & 73.7 & 32.5  & 35.4 & 3.9   & 30.1      & 28.1 & 42.4 \\
%
% CyCADA \cite{hoffman2018cycada}  
% & 86.7 & 35.6    & 80.1     & 19.8 & 17.5  & 38.0 & 39.9  & 41.5 & 82.7       & 27.9    & 73.6 & 64.9   & 19.0  & 65.0 & 12.0  & 28.6 & 4.5   & 31.1      & 42.0 & 42.7 \\

% CLAN \cite{luo2019taking} & 87.0 & 27.1 & 79.6 & 27.3 & 23.3 & 28.3 & 35.5 & 24.2 & 83.6 & 27.4 & 74.2 & 58.6 & 28.0 & 76.2 & 33.1 & 36.7 & 6.7 & 31.9 & 31.4 & 43.2 \\
% %
% ADVENT \cite{vu2019advent}
% & 89.4 & 33.1    & 81.0     & 26.6 & 26.8  & 27.2 & 33.5  & 24.7 & 83.9       & 36.7    & 78.8 & 58.7   & 30.5  & 84.8 & 38.5  & 44.5 & 1.7   & 31.6      & 32.4 & 45.5 \\
% %
% CBST \cite{zou2018unsupervised} & 91.8 & 53.5 & 80.5 & 32.7 & 21.0 & 34.0 & 28.9 & 20.4 & 83.9 & 34.2 & 80.9 & 53.1 & 24.0 & 82.7 & 30.3 & 35.9 & 16.0 & 25.9 & 42.8 & 45.9\\
% %
FADA \cite{wang2020classes}
& 92.5 & 47.5    & 85.1     & 37.6 & 32.8  & 33.4 & 33.8  & 18.4 & 85.3       & 37.7    & 83.5 & 63.2   & 39.7  & 87.5 & 32.9  & 47.8 & 1.6   & 34.9      & 39.5 & 49.2 \\
Uncertainty \cite{zheng2021rectifying} & 90.4 & 31.2 & 85.1 & 36.9 & 25.6 & 37.5 & 48.8 & 48.5 & 85.3 & 34.8 & 81.1 & 64.4 & 36.8 & 86.3 & 34.9 & 52.2 & 1.7 & 29.0 & 44.6 & 50.3 \\
FDA \cite{yang2020fda}
& 92.5 & 53.3    & 82.4     & 26.5 & 27.6  & 36.4 & 40.6  & 38.9 & 82.3       & 39.8    & 78.0 & 62.6   & 34.4  & 84.9 & 34.1  & 53.1 & 16.9  & 27.7      & 46.4 & 50.5 \\
DACS \cite{tranheden2021dacs} & 89.9 & 39.7 & 87.9 & 30.7 & 39.5 & 38.5 & 46.4 & 52.8 & 88.0 & 44.0 & 88.8 & 67.2 & 35.8 & 84.5 & 45.7 & 50.2 & 0.0 & 27.3 & 34.0 & 52.1\\
CorDA \cite{wang2021domain} & 94.7 & 63.1 & 87.6 & 30.7 & 40.6 & 40.2 & 47.8 & 51.6 & 87.6 & 47.0 & 89.7 & 66.7 & 35.9 & 90.2 & 48.9 & 57.5 & 0.0 & 39.8 & 56.0 & 56.6\\
ProDA \cite{zhang2021prototypical} & 87.8 & 56.0 & 79.7 & 46.3 & 44.8 & 45.6 & 53.5 & 53.5 & 88.6 & 45.2 & 82.1 & 70.7 & 39.2 & 88.8 & 45.5 & 59.4 & 1.0 & 48.9 & 56.4 & 57.5\\
CONFETI \cite{li2023confeti}     & 96.5 & 75.6 & 88.9 & 45.1 & 45.9 & 50.1 & \textbf{61.2} & \textbf{68.2} & 89.4 & 45.7 & 86.3 & \textbf{76.3} & \textbf{49.9} & 92.2 & 55.1 & 62.8 & 16.7 & 33.8 & 63.1 & 63.3\\ 
IDM \cite{wang2023informative}  & \textbf{97.2} & \textbf{77.1} & 89.8 & 51.7 & \underline{51.7} & \underline{54.5} & 59.7 & 64.7 & 89.2 & 45.3 & 90.5 & \underline{74.2} & 46.6 & 92.3 & 76.9 & 59.6 & \underline{81.2} & 57.3 & 62.4 & 69.5\\
\hline
DAFormer \cite{hoyer2022daformer} & 95.7 & 70.2 & 89.4 & 53.5 & 48.1 & 49.6 & 55.8 & 59.4 & 89.9 & 47.9 & \underline{92.5} & 72.2 & 44.7 & 92.3 & 74.5 & 78.2 & 65.1 & 55.9 & 61.8 & 68.3\\
DAFormer+FST \cite{du2022learning} & 95.3 & 67.7 & 89.3 & \underline{55.5} & 47.1 & 50.1 & 57.2 & 58.6 & 89.9 & 51.0 & \textbf{92.9} & 72.7 & 46.3 & 92.5 & 78.0 & 81.6 & 74.4 & \textbf{57.7} & 62.6 & 69.3 \\ 
DAFormer+CAMix \cite{zhou2022context}  & 96.0 & 73.1 & 89.5 & 53.9 & 50.8 & 51.7 & 58.7 & 64.9 & 90.0 & \underline{51.2} &92.2 &71.8 &44.0 & 92.8 & \textbf{78.7} & \underline{82.3} &70.9 &54.1 & \underline{64.3} & 70.0 \\
DAFormer+MICDrop \cite{yang2024micdrop}  & 96.0 & 71.8 & \underline{90.3} & 53.3 & 46.4 & \textbf{54.8} &57.8 &\underline{66.7} & \underline{90.0} &49.2 &92.2 &73.6 &46.3 &\underline{92.8} &\underline{78.1} &80.6 &70.7 &\underline{57.5} &63.2 & \underline{70.1} \\
\textbf{DAFormer+OMUDA} & \underline{96.9} & \underline{76.6} & \textbf{90.8} & \textbf{59.8} & \textbf{52.3} & 54.1 & \underline{59.7} & 64.9 & \textbf{90.9} & \textbf{51.7} & 92.3 & 73.1 & \underline{49.3} & \textbf{93.5} & 77.7 & \textbf{82.9} & \textbf{81.6} & 54.6 & \textbf{64.9} & \textbf{72.0}\\
\toprule
\end{tabular}
}
\end{table*}

\begin{table*}[t]
\centering
\caption{Quantitative comparison with previous UDA methods on SYNTHIA$\to$Cityscapes benchmark. The best results are highlighted in bold while the second best results are shown by underline.}
\label{tab2}
\setlength{\tabcolsep}{1pt}
\renewcommand{\arraystretch}{1}
\resizebox{\textwidth}{!}{%
\begin{tabular}{l|ccccccccccccccccccc|c}
\bottomrule
 & \rotatebox{45}{Road} & \rotatebox{45}{S.walk} & \rotatebox{45}{Build.} & \rotatebox{45}{Wall} & \rotatebox{45}{Fence} & \rotatebox{45}{Pole} & \rotatebox{45}{Tr.Light} & \rotatebox{45}{Sign} & \rotatebox{45}{Veget.} & \rotatebox{45}{Terrain} & \rotatebox{45}{Sky} & \rotatebox{45}{Person} & \rotatebox{45}{Rider} & \rotatebox{45}{Car} & \rotatebox{45}{Truck} & \rotatebox{45}{Bus} & \rotatebox{45}{Train} & \rotatebox{45}{M.bike} & \rotatebox{45}{Bike} & mIoU\\
\hline
\multicolumn{21}{c}{SYNTHIA $\to$ Cityscapes} \\
\hline
% ADVENT \cite{vu2019advent}
% & 85.6 & 42.2    & 79.7     & 8.7  & 0.4   & 25.9 & 5.4   & 8.1  & 80.4 & -- & 84.1 & 57.9   & 23.8  & 73.3 & -- & 36.4 & -- & 14.2 & 33.0 & 41.2 \\
% CBST \cite{zou2018unsupervised} & 68.0 & 29.9 & 76.3 & 10.8 & 1.4 & 33.9 & 22.8 & 29.5 & 77.6 & -- & 78.3 & 60.6 & 28.3 & 81.6 & -- & 23.5 & -- & 18.8 & 39.8 & 42.6\\
% MRNet \cite{zheng2019unsupervised} & 82.0 & 36.5 & 80.4 & 4.2 & 0.4 & 33.7 & 18.0 & 13.4 & 81.1 & -- & 80.8 & 61.3 & 21.7 & 84.4 & -- & 32.4 & -- & 14.8 & 45.7  & 43.2 \\
% %
FADA \cite{wang2020classes} 
& 84.5 & 40.1    & 83.1     & 4.8  & 0.0   & 34.3 & 20.1  & 27.2 & 84.8 & -- & 84.0 & 53.5   & 22.6  & 85.4 & -- & 43.7 & -- & 26.8 & 27.8 & 45.2 \\
Uncertainty \cite{zheng2021rectifying} & 87.6 & 41.9 & 83.1 & 14.7 & 1.7 & 36.2 & 31.3 & 19.9 & 81.6 & --  & 80.6 & 63.0 & 21.8 & 86.2 & -- & 40.7 & -- & 23.6 & 53.1  & 47.9 \\
CCM \cite{li2020content} & 79.6 & 36.4 & 80.6 & 13.3 & 0.3 & 25.5 & 22.4 & 14.9 & 81.8 & -- & 77.4 & 56.8 & 25.9 & 80.7 & -- & 45.3 & -- & 29.9 &  52.0  & 45.2 \\
DACS \cite{tranheden2021dacs} & 80.6 & 25.1 & 81.9 & 21.5 & 2.9 & 37.2 & 22.7 & 24.0 & 83.7 & -- & 90.8 & 67.6 & 38.3 & 82.9 & -- & 38.9 & -- & 28.5 & 47.6 & 48.3\\
CorDA \cite{wang2021domain} & \textbf{93.3} & \textbf{61.6} & 85.3 & 19.6 & 5.1 & 37.8 & 36.6 & 42.8 & 84.9 & -- & 90.4 & 69.7 & 41.8 & 85.6 & -- & 38.4 & -- & 32.6 & 53.9 & 55.0\\
ProDA \cite{zhang2021prototypical} & 87.8 & 45.7 & 84.6 & 37.1 & 0.6 & 44.0 & 54.6 & 37.0 & \textbf{88.1} & -- & 84.4 & \underline{74.2} & 24.3 & 88.2 & -- & 51.1 & -- & 40.5 & 45.6 & 55.5\\

CONFETI \cite{li2023confeti}     & 83.8 & 44.6 & 86.9 & 15.4 & 3.7 & 44.3 & 56.9 & 55.5 & 84.9 & -- & 86.2 & 73.8 & 46.8 & \textbf{90.1} & -- & 57.1 & -- & 46.0 & 63.2 & 58.7\\ 
IDM \cite{wang2023informative}   & 87.6 & 47.6 & 88.1 & 33.4 & 6.3 & \textbf{52.8} & \underline{57.8} & 56.5 & 83.0 & -- & 77.5 & 66.2 & \underline{52.1} & \underline{89.3} & -- & 55.6 & -- & \underline{57.1} & \textbf{64.2} & 60.9\\
\hline
DAFormer \cite{hoyer2022daformer} & 84.5 & 40.7 & 88.4 & 41.5 & 6.5 & 50.0 & 55.0 & 54.6 & 86.0 & -- & 89.8 & 73.2 & 48.2 & 87.2 & -- & 53.2 & -- & 53.9 & 61.7 & 60.9\\

DAFormer+FST \cite{du2022learning} & 88.3 & 46.1 & 88.0 & 41.7 & 7.3 & 50.1 & 53.6 & 52.5 & \underline{87.4} & -- & \underline{91.5} & 73.9 & 48.1 & 85.3 & -- & 58.6 & -- & 55.9 & \underline{63.4} & 61.9 \\

DAFormer+MICDrop \cite{yang2024micdrop} & 81.0 &37.1 &\underline{89.4} & \underline{45.7} & \textbf{9.5} &51.8 &57.3 & \underline{58.0} &86.7 &-- &85.0 &73.6 &50.4 &88.2 &-- &\textbf{64.7} &-- &56.8 &62.8 & \underline{62.4} \\

\textbf{DAFormer+OMUDA} & \underline{88.5} & \underline{51.9} & \textbf{89.6} & \textbf{49.8} & \underline{7.9} & \underline{52.4} & \textbf{59.2} & \textbf{58.6} & 86.2 & -- & \textbf{92.8} & \textbf{76.2} & \textbf{54.0} & 88.9 & -- & \underline{64.2} & -- & \textbf{57.9} & 61.8 & \textbf{65.0}\\
\toprule
\end{tabular}
}
\end{table*}

\begin{table}[!t]
\caption{Quantitative comparison with previous UDA methods on GTA5$\to$Cityscapes and SYNTHIA$\to$Cityscapes benchmark. The best mIoU results are highlighted in bold while the second best results are shown by underline.}
\label{tab3}
\centering
\setlength{\tabcolsep}{8.5pt}
\renewcommand{\arraystretch}{1.0}
\scriptsize
\begin{tabular}{c|c|c}
\bottomrule
   & {GTA5 $\to$ Cityscapes} & {SYNTHIA $\to$ Cityscapes} \\ \hline
VLTSeg \cite{hummer2023vltseg} & 62.6 & 56.8  \\ 
TQDM   \cite{pak2024textual}    & 68.9 & 58.0  \\
DPMFormer \cite{jeon2025exploiting} & \underline{70.1} & \underline{58.9}  \\
\textbf{DAFormer+OMUDA}            & \textbf{72.0} & \textbf{65.0}   \\ \toprule
\end{tabular}
\end{table}

\begin{figure*}[t]
\centering
\includegraphics[width=\linewidth]{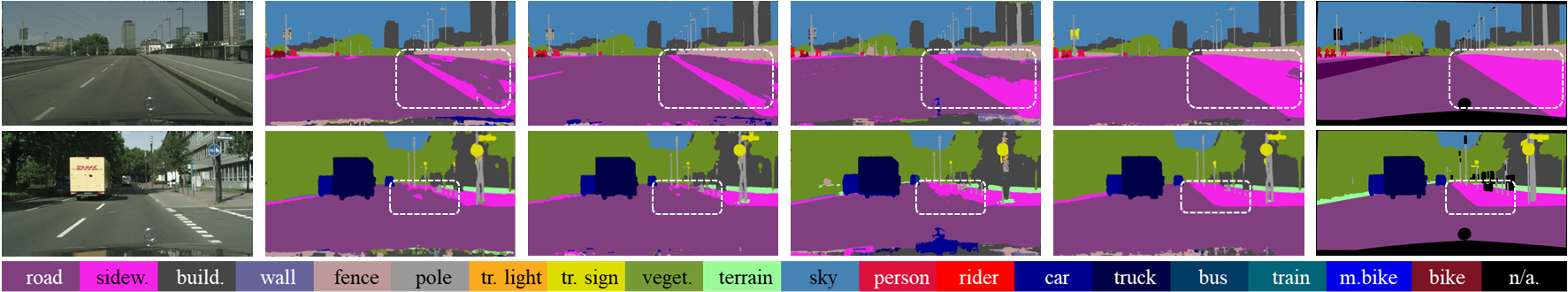}\\
\caption{Qualitative results on GTA5$\to$Cityscapes. From left to right: Target image, the visual results predicted by DAFormer, DAFormer+FST, DPMFormer, DAFormer+OMUDA (ours), and Ground Truth. We deploy the white rectangles highlight the visual improvements. OMUDA  better segments difficult classes, including sidewalk, terrain, traffic sign, and bus.}
\label{fig3}
\end{figure*}

\begin{figure*}[t]
\centering
\includegraphics[width=\linewidth]{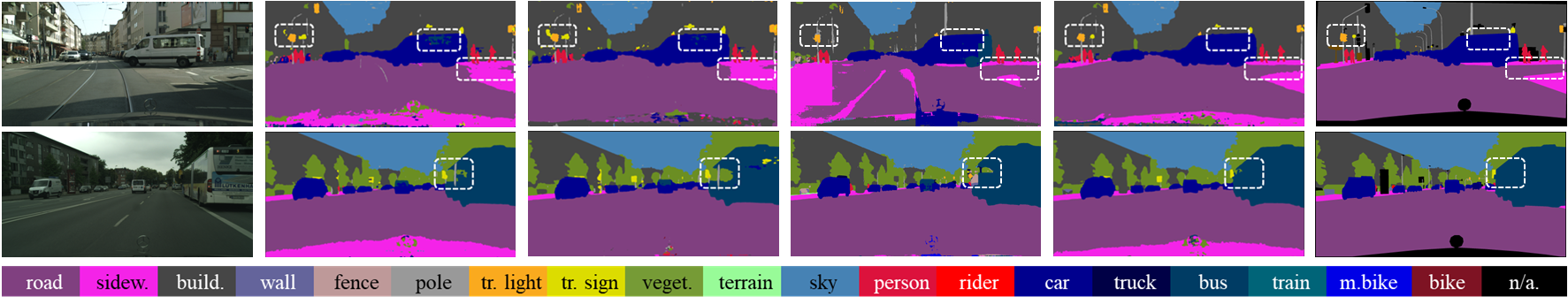}\\
\caption{Qualitative results on SYNTHIA$\to$Cityscapes. From left to right: Target Image, the visual results predicted by DAFormer, DAFormer+FST, DPMFormer, DAFormer+OMUDA (Ours), and Ground Truth. We deploy the white rectangles highlight the visual improvements. OMUDA better segments difficult classes such as sidewalk, terrain, traffic sign, and bus.}
\label{fig4}
\vspace{-1.0em}
\end{figure*}

\begin{figure}[t]
\centering
\includegraphics[width=0.99\linewidth]{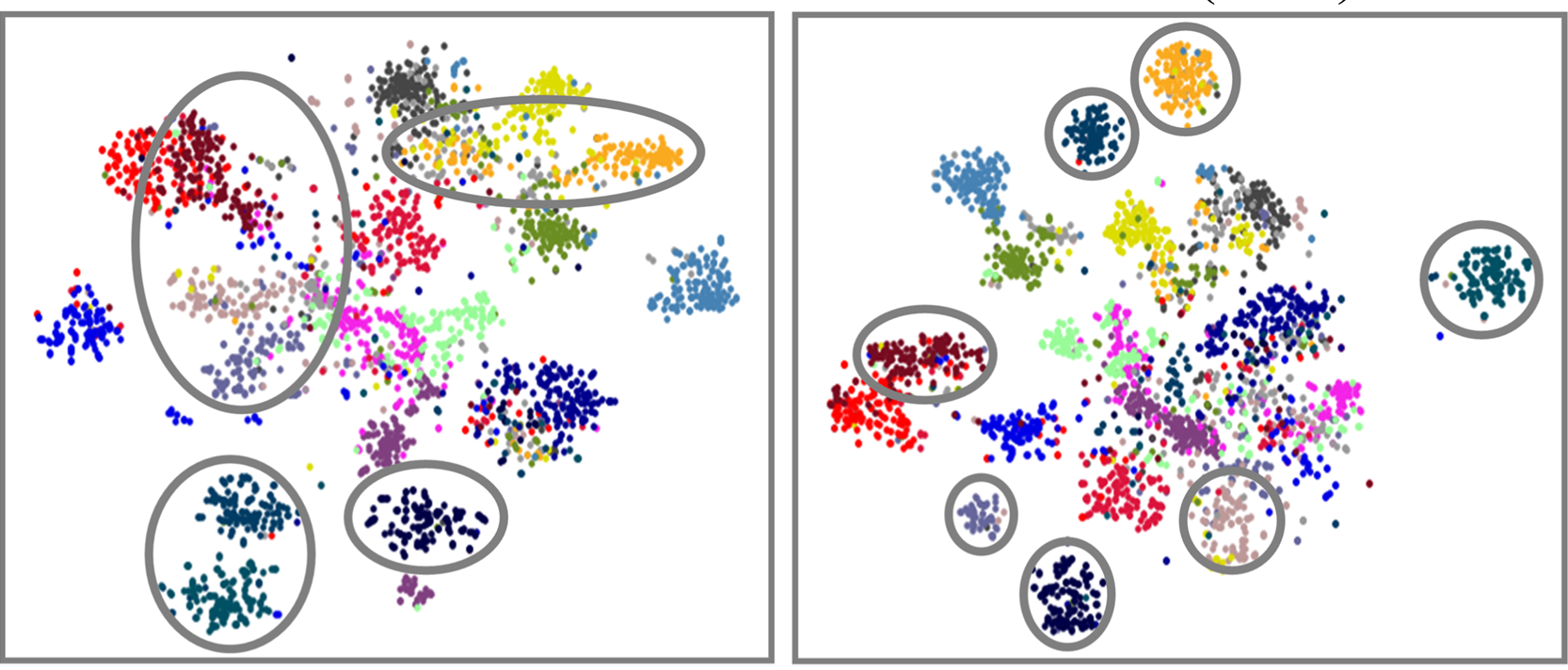}
\caption{T-SNE visualization of learned feature embeddings for DAFormer (left) and OMUDA (right) on Cityscapes dataset.}
\label{fig:tsne_imagenet}
\end{figure}

\subsection{Comparison with the State-of-the-art Approaches}
\label{Section IV.B}
In this section, we compare the performance of the proposed OMUDA with those of the state-of-the-art UDA methods on GTA5$\to$Cityscapes and SYNTHIA$\to$Cityscapes, respectively. Specifically, we use FADA (ECCV'2020) \cite{wang2020classes}, Uncertainty (IJCV'2021) \cite{zheng2021rectifying}, FDA (CVPR'2020) \cite{yang2020fda}, DACS (IWCACV'2021) \cite{tranheden2021dacs}, CorDA (CVPR'2021) \cite{wang2021domain}, ProDA (CVPR'2021) \cite{zhang2021prototypical}, DAFormer(2022'CVPR) \cite{hoyer2022daformer}, DAFormer+FST (NIPS'2022) \cite{du2022learning}, DAFormer+CAMix (TCSVT'2022) \cite{zhou2022context}, CONFETI (CVPR'2023) \cite{li2023confeti}, IDM (CVPR'2023) \cite{wang2023informative}, VLTSeg (arXiv'2023) \cite{hummer2023vltseg},  DAFormer+MICDrop (ECCV'2024) \cite{yang2024micdrop}, TQDM (ECCV'2024) \cite{pak2024textual}, DPMFormer (CVPR'2025) \cite{jeon2025exploiting}. Quantitative comparisons are shown in Table \ref{tab1} to Table \ref{tab3}. 
\subsubsection{GTA5$\to$Cityscapes} First, we present our results on GTA$\to$Cityscapes in Table \ref{tab1} and highlight the best results in bold. Generally, our OMUDA achieves 72.0 mIoU, yielding a significant improvement over the previous strongest transformer-based models, including DAFormer\cite{hoyer2022daformer}, DAFormer+FST\cite{du2022learning}, DAFormer+CAMix\cite{zhou2022context} and DAFormer+MICDrop \cite{yang2024micdrop}. Among all the 19 classes, we achieve the best scores on 9 classes and the second best scores on 4 classes including several hard objectives such as wall, fence, terrian, and train, which cannot be well handled in previous works. Particularly, we increase the IoU of the train by +7.2 from 74.4 to 81.6 mIoU. The IoU performance of OMUDA verifies our motivation that class decoupling indeed helps category recognition, especially for challenging small or rare objects.
\subsubsection{SYNTHIA$\to$Cityscapes} Table \ref{tab2} displays the adaptation results on SYNTHIA $\to$ Cityscapes, demonstrating remarkable advancements achieved by OMUDA. The proposed OMUDA increases +4.1, +3.1, +2.6 mIoU compared with DAFormer\cite{hoyer2022daformer}, DAFormer+FST\cite{du2022learning} and DAFormer+MICDrop\cite{yang2024micdrop}, respectively. Among all the 16 classes, we achieve the best scores on 8 classes and the second best scores on 5 classes, most of those are hard classes that previous approaches struggled to effectively address, e.g., wall, traffic light, person and rider.  We have also report the comparison results of our ODMA method with VLTSeg\cite{hummer2023vltseg}, TQDM\cite{pak2024textual}, and DPMFormer\cite{jeon2025exploiting} in Table \ref{tab3}. 
\subsubsection{Qualitative results} Additionally, we provide qualitative results of our approach in comparison with DAFormer\cite{hoyer2022daformer}, DAFormer+FST\cite{du2022learning} and DPMFormer \cite{jeon2025exploiting}, as depicted in Figs. \ref{fig3} and \ref{fig4}. In these two Figs, we visualize the segmentation results and the comparison with previous strong methods DAFormer\cite{hoyer2022daformer}, DAFormer+FST\cite{du2022learning} and DPMFformer \cite{jeon2025exploiting}, and the ground truth on GTA5$\to$Cityscapes and SYNTHIA$\to$Cityscapes benchmarka. The results highlighted by white dash boxes show that OMUDA corrects typical classes that cannot be well handled DAFormer, DAFormer+FST and DPMFformer, e.g. sidewalk, fence, and bus. Especially in the second row of Fig. \ref{fig4}, we can see that OMUDA is more complete and smoother in terms of object interiors and edge details. This outcome is attributed to class decoupling, wherein each class is trained based on its distinctive attributes. This strategy effectively alleviates class confusion and yields smoother edge predictions.

% \begin{table*}[t]
% \centering
% \caption{\textcolor{red}{Comparison of the class-wise IoU of each proposed component for GTA$\to$Cityscapes.}}
% \label{tab:ab}
% \setlength{\tabcolsep}{7pt}
% \renewcommand{\arraystretch}{1.2}
% \small
    
% \begin{tabular}{l|cccccccccc}
% \bottomrule
% & \rotatebox{45}{Road} & \rotatebox{45}{S.walk} & \rotatebox{45}{Build.} & \rotatebox{45}{Wall} & \rotatebox{45}{Fence} & \rotatebox{45}{Pole} & \rotatebox{45}{T.Light} & \rotatebox{45}{T.Sign} & \rotatebox{45}{Veget.} & \rotatebox{45}{Terrain} \\
% \hline
% Baseline & 95.7 & 70.2 & 89.4 & 53.5 & 48.1 & 49.6 & 55.8 & 59.4 & 89.9 & 47.9 \\

% +CAM & 96.6 & 74.7 & 89.3 & 54.1 & 45.5 & 47.8 & 56.9 & 58.4 & 89.9 & 50.9 \\
% +FDM & 96.0 & 71.9 & 89.7 & 55.4 & 49.1 & 49.8 & 56.0 & 62.2 & 89.9 & 49.8 \\
% +CDM & 97.0 & 76.3 & 90.3 & 56.7 & 53.4 & 53.5 & 58.5 & 60.0 & 90.7 & 51.0 \\
% All & 96.9 & 76.6 & 90.8 & 59.8 & 52.3 & 54.1 & 59.6 & 64.9 & 90.9 & 51.7 \\
% \hline
% & \rotatebox{45}{Sky} & \rotatebox{45}{Person} & \rotatebox{45}{Rider} & \rotatebox{45}{Car} & \rotatebox{45}{Truck} & \rotatebox{45}{Bus} & \rotatebox{45}{Train} & \rotatebox{45}{M.bike} & \rotatebox{45}{Bike} & \\
% \hline
% Baseline & 92.5 & 72.2 & 44.7 & 92.3 & 74.5 & 78.2 & 65.1 & 55.9 & 61.8 & \\
% +CAM & 90.8 & 71.9 & 43.7 & 92.4 & 78.7 & 78.8 & 69.7 & 57.7 & 63.7 & \\
% +FDM & 90.3 & 72.8 & 42.6 & 92.5 & 78.8 & 79.1 & 71.6 & 57.7 & 63.4 & \\
% +CDM & 92.1 & 73.1 & 48.5 & 93.2 & 76.7 & 82.4 & 79.3 & 54.7 & 64.6 & \\
% All & 92.3 & 73.1 & 49.3 & 93.5 & 77.7 & 82.9 & 81.6 & 54.6 & 64.9 & \\
% \toprule

% \end{tabular}
% \end{table*}

\begin{table*}[t]
\centering
\caption{Comparison of the class-wise IoU of each proposed component for GTA$\to$Cityscapes.}
\label{tab4}
\setlength{\tabcolsep}{1pt}
\renewcommand{\arraystretch}{1}
\resizebox{\textwidth}{!}{%
\begin{tabular}{l|ccccccccccccccccccc|c}
\bottomrule
 & \rotatebox{45}{Road} & \rotatebox{45}{S.walk} & \rotatebox{45}{Build.} & \rotatebox{45}{Wall} & \rotatebox{45}{Fence} & \rotatebox{45}{Pole} & \rotatebox{45}{Tr.Light} & \rotatebox{45}{Sign} & \rotatebox{45}{Veget.} & \rotatebox{45}{Terrain} & \rotatebox{45}{Sky} & \rotatebox{45}{Person} & \rotatebox{45}{Rider} & \rotatebox{45}{Car} & \rotatebox{45}{Truck} & \rotatebox{45}{Bus} & \rotatebox{45}{Train} & \rotatebox{45}{M.bike} & \rotatebox{45}{Bike} & mIoU\\
\hline
Baseline & 95.7 & 70.2 & 89.4 & 53.5 & 48.1 & 49.6 & 55.8 & 59.4 & 89.9 & 47.9 & \textbf{92.5} & 72.2 & 44.7 & 92.3 & 74.5 & 78.2 & 65.1 & 55.9 & 61.8 & 68.3\\
+CAM & 96.6 & 74.7 & 89.3 & 54.1 & 45.5 & 47.8 & 56.9 & 58.4 & 89.9 & 50.9 & 90.8 & 71.9 & 43.7 & 92.4 & 78.7 & 78.8 & 69.7 & \textbf{57.7} & 63.7 & 69.3\\
+FDM & 96.0 & 71.9 & 89.7 & 55.4 & 49.1 & 49.8 & 56.0 & 62.2 & 89.9 & 49.8 & 90.3 & 72.8 & 42.6 & 92.5 & \textbf{78.8} & 79.1 & 71.6 & 57.7 & 63.4 & 69.1\\
+CDM & \textbf{97.0} & 76.3 & 90.3 & 56.7 & \textbf{53.4} & 53.5 & 58.5 & 60.0 & 90.7 & 51.0 & 92.1 & 73.1 & 48.5 & 93.2 & 76.7 & 82.4 & 79.3 & 54.7 & 64.6 & 71.2\\
All & 96.9 & \textbf{76.6} & \textbf{90.8} & \textbf{59.8} & 52.3 & \textbf{54.1} & \textbf{59.6} & \textbf{64.9} & \textbf{90.9} & \textbf{51.7} & 92.3 & \textbf{73.1} & \textbf{49.3} & \textbf{93.5} & 77.7 & \textbf{82.9} & \textbf{81.6} & 54.6 & \textbf{64.9} & \textbf{72.0}\\
\toprule
\end{tabular}
}
\end{table*}

% \begin{table}[t]
% \centering
% \caption{Ablation studies of each proposed component for GTA$\to$Cityscapes.}
% \label{tab:ablation}
% \setlength{\tabcolsep}{3pt}
% \renewcommand{\arraystretch}{1.2}
% \small
% \begin{tabular}{lcccc}
% \bottomrule
%  & \makecell{Pseudo-label\\Masking} 
%  & \makecell{Context-based\\Masking} 
%  & \makecell{Feature\\Masking} 
%  & mIoU \\
% \hline\hline
% 1 &   &   &   & 68.3 \\
% 2 & $\surd$ &  &  & 69.3 \\
% 3 & & $\surd$ &  & 69.1 \\
% 4 & & & $\surd$ & 71.2  \\
% 5 & $\surd$  & $\surd$ & $\surd$ & 72.0 \\
% \hline
% \toprule
% \end{tabular}
% \end{table}

To better develop intuition, we draw T-SNE visualizations\cite{van2008visualizing} of the learned feature representations for DAFormer and OMUDA in Fig. \ref{fig:tsne_imagenet}. From the T-SNE analysis, we can observe that although DAFormer can produce separated features, they are mainly focused on simple classes (e.g., sky) and the same class may produce multiple clusters of features (e.g., traffic light), while the interval between similar classes is not obvious (e.g., fence and wall), making it difficult to make dense predictions. With the implementation of class decoupling, features among different classes can achieve better intra-class compactness and inter-class separability, especially for classes that are usually particularly difficult to adapt, such as traffic light, fence, wall, truck, bus, and train. In contrast, the embedded representations of our OMUDA showcase the most distinct clusters when compared to those of DAFormer. This observation underscores the discriminative potential underpinning the class decoupling approach.

\subsection{Ablation Study}
\label{Section IV.C}
In this section, a series of ablation studies are conducted to substantiate the individual contributions of each component to the ultimate performance. All experiments are performed on the GTA5$\rightarrow$Cityscapes benchmark.
\subsubsection{Effect of Components}
We conduct a comprehensive series of experiments to assess the efficacy of the developed OMUDA framework and report the performance in Table \ref{tab4}. We take DAFormer\cite{hoyer2022daformer} as our baseline. Based on the results, we note that each component significantly contributes to the overall performance of our framework. Notably, the inclusion of the CDM component yields the most substantial gain of +2.9 mIoU. Specifically, it can be seen from Table \ref{tab4} that adding the CAM sharply improves performance for difficult-to-adapt classes, such as sidewalk, terrain, truck, and train. And with the addition of the FDM, the performance improvement mainly comes from rare classes, especially in fence, pole, traffic signal, and train.

\begin{table}[t]
\centering
\caption{The effect of loss weight on GTA5$\to$Cityscapes.}
\label{tab:sampling}
\setlength{\tabcolsep}{12pt}
\renewcommand{\arraystretch}{1.2}
\small
\begin{tabular}{ c | c c c c} 
\bottomrule
Strategy & Uniform & Fix radio & RCS & Ours \\ 
\hline 
mIoU  & 69.8 & 71.5 &  71.6 & {\bf 72.0} \\ 
\toprule
\end{tabular}
\end{table}

\subsubsection{Ablation Study on CAM}
We compare several data sampling strategies, including uniform sampling, a fixed sampling ratio, RCS (as adopted in DAFormer\cite{hoyer2022daformer}), and our proposed CAM strategy. Table~\ref{tab:sampling} summarizes the results. Across all evaluations, our approach consistently delivers superior performance, demonstrating its effectiveness over conventional sampling methods. Furthermore, we evaluate the effectiveness of our masking strategy by comparing it with random masking\cite{xie2022simmim} and grid masking\cite{chen2020gridmask}. As shown in Table~\ref{tab:mask2}, CAM consistently outperforms both baselines, achieving gains of +0.6 mIoU over random masking and +0.8 mIoU over grid masking. Notably, different CAM configurations also impact performance: a finer-grained setting (16/32) yields the best result, reaching 72.0 mIoU. These findings demonstrate that context-aware masking not only surpasses conventional masking strategies but also benefits from carefully chosen mask granularity.

\subsubsection{Effect of the KD Loss Weight in the FDM}
Table \ref{tab:KD2} reports the results obtained with different KD loss weights ($\lambda_{KD}\in {0.1, 0.05, 0.01, 0.005}$). These results highlight the critical role of the KD loss in stabilizing feature learning within the FDM module. When the weight is too large, the student model focuses excessively on matching the pretrained features, which limits its ability to learn target-domain representations and leads to reduced adaptation performance. In contrast, if the weight is too small, the distillation signal becomes insufficient to regularize the model, making it prone to overfitting the target pseudo labels and producing unstable features. These observations indicate that an appropriate KD weight strikes the right balance by providing effective regularization while preserving the model’s capacity to adapt, thus yielding the best performance.

\begin{table}[t]
\centering
\caption{The influence of different masking strategy on GTA5$\to$Cityscapes.}
\label{tab:mask2}
\setlength{\tabcolsep}{8pt}
\renewcommand{\arraystretch}{1.2}
\small
\begin{tabular}{cccc}
    \bottomrule
    Mask strategy & Fore-mask size & Back-mask size &  mIoU \\
    \hline
    Random & -- & -- &  70.6\\
    Grid & -- & -- &  70.4\\
    CAM & 32 & 64  &  70.9\\
   CAM & 16 & 32  &  {\bf 72.0}\\
    \toprule
\end{tabular}
\end{table}

\begin{table}[t]
\centering
\caption{The effect of KD Loss Weight on GTA5$\to$Cityscapes.}
\label{tab:KD2}
\setlength{\tabcolsep}{16pt}
\renewcommand{\arraystretch}{1.2}
\small
\begin{tabular}{l|ccccc} 
\bottomrule
$\lambda_{KD}$ & 0.1 & 0.05 & 0.01 & 0.005 \\ 
\hline 
mIoU & 69.4 & 70.2 & {\bf 72.0} & 70.1 \\ 
\toprule
\end{tabular}
\end{table}

\section{Conclusion and Future Work}
\label{Section V}
\subsection{Conclusion}
\label{Section V.A}
In this paper, we propose OMUDA, a unified masking-based framework to mitigate performance degradation in semantic segmentation under domain shift. OMUDA integrates context-aware masking for intelligent data mixing, feature distillation to promote domain-invariant representations, and category-specific masking to stabilize pseudo-label supervision and enhance feature compactness. By jointly leveraging these complementary mechanisms, OMUDA effectively strengthens cross-domain knowledge transfer. Notably, OMUDA is implemented purely as a training strategy and introduces no additional computational overhead at inference. We believe OMUDA provides useful insights for future research in unsupervised domain adaptation and the design of more robust semantic segmentation models.

\subsection{Limitations and Future Work}
\label{Section V.B}
Our framework requires defining foreground–background class groups and tuning several hyperparameters, though these requirements are common in UDA research. We have provided detailed analyses to help guide these choices. Future directions include developing adaptive masking strategies that dynamically respond to domain shift severity, incorporating self-supervised objectives to further enhance representation robustness, and extending the approach to multi-source domain adaptation settings.


% \section*{Declaration of competing interest}
% The authors declare that they have no known competing financial interests or personal relationships that could have appeared to influence the work reported in this paper.

% \section*{Acknowledgment}
% This study was supported by the Natural Science Foundation of Qinghai Province under Grant 2023-QLGKLYCZX-017.

\bibliographystyle{IEEEtran}

\bibliography{IEEEreference}

\end{document}